\documentclass{article}
\usepackage{graphicx}
\usepackage{amsmath}
\usepackage{hyperref}
\usepackage{amsfonts}
\usepackage{tikz}
\usetikzlibrary{shapes.geometric, arrows, positioning}
\usepackage{amssymb}

\usepackage[english]{babel}

\usepackage{graphicx}
\usepackage{subfig}

\tikzstyle{block} = [rectangle, rounded corners, minimum width=3cm, minimum height=1cm, text centered, draw=black, fill=blue!20]
\tikzstyle{process} = [rectangle, minimum width=3cm, minimum height=1cm, text centered, draw=black, fill=green!30]
\tikzstyle{arrow} = [thick,->,>=stealth]
\tikzstyle{decision} = [diamond, draw, fill=orange!30, text centered, inner sep=0.1cm]

\usepackage[final,nonatbib]{neurips_2024}

\title{A Tutorial on LLM Reasoning:\\
Relevant Methods behind ChatGPT o1}
\author{Jun Wang\\jun.wang@cs.ucl.ac.uk\\
UCL Centre for Artificial Intelligence}

\begin{document}

\maketitle

\begin{abstract}
OpenAI o1 has shown that applying reinforcement learning to integrate reasoning steps directly during inference can significantly improve a model's reasoning capabilities. This result is exciting as the field transitions from the conventional autoregressive method of generating answers to a more deliberate approach that models the slow-thinking process through step-by-step reasoning training. Reinforcement learning plays a key role in both the model's training and decoding processes. In this article, we present a comprehensive formulation of reasoning problems and investigate the use of both model-based and model-free approaches to better support this slow-thinking framework.
\end{abstract}

\begin{figure*}[h]
    \centering
    \includegraphics[width=1\linewidth]{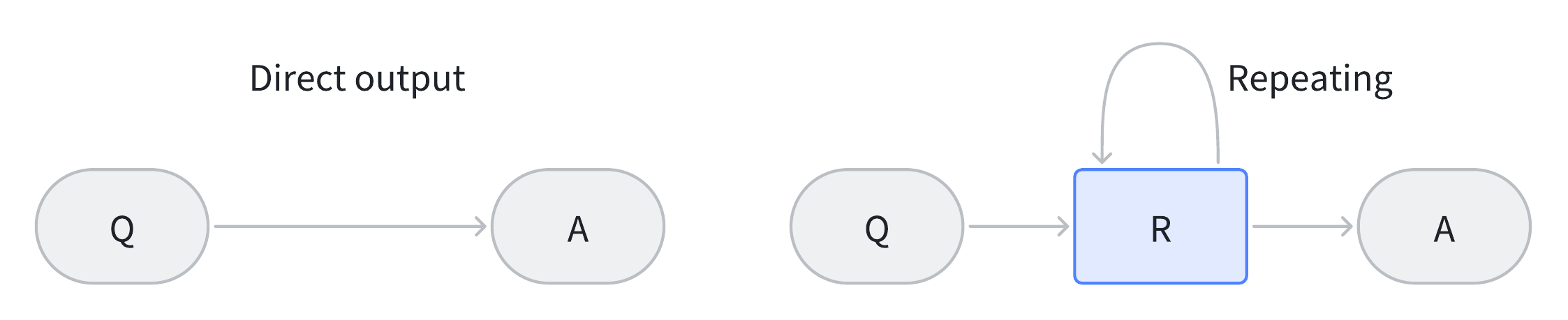}
    (a)  \qquad\qquad\qquad\qquad\qquad\qquad\qquad\qquad \qquad\qquad  (b)
    \caption{Inference-time computation. (a) An autoregressive LLM directly generate an answer (A) by conditioning on the given question (Q).
    (b) The concept of chain of thought, or step-by-step thinking, involves incorporating intermediate reasoning steps (R) before arriving at the final answer (A). These repeated operations allow for 1) revisiting and revising prior outputs, 2) progressing to subsequent reasoning stages, and 3) exploring multiple reasoning paths or trajectories.}
    \label{fig:cot}
\end{figure*}

\section{Background}
OpenAI has recently unveiled ChatGPT o1 \cite{o1}, a groundbreaking Large Language Model (LLM) that represents a giant leap forward in strong AI. 
Trained using reinforcement learning techniques, o1 excels in complex reasoning tasks by explicitly embedding a \emph{native} ``Chain-of-Thought'' (NCoT) process, which allows it to ``deep think" through step-by-step reasoning before generating responses. The model is reported to be five times more proficient in math and coding compared to the previous ChatGPT 4o, specifically displaying exceptional performance across various domains: it ranks in the 89th percentile for competitive programming, places among the top 500 students in a prestigious US math olympiad qualifier, and surpasses human PhD-level accuracy in physics, biology, and chemistry benchmarks. A key innovation of o1 is that it allows spending more time reasoning during the inference process, marking a shift from fast, direct responses to slow, deliberate, multi-step inference-time computation (Fig.~\ref{fig:cot}).

\begin{figure}[t]
    \centering
    \subfloat[\centering System 1 nonconscious control.]{{\includegraphics[width=6.5cm]{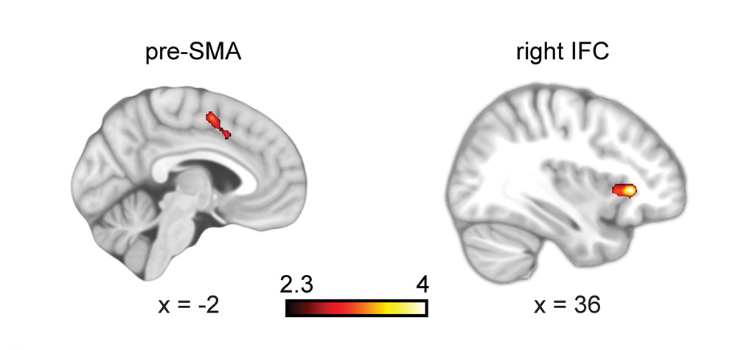} }}%
    \qquad
    \subfloat[\centering System 2 conscious control.]{{\includegraphics[width=6.5cm]{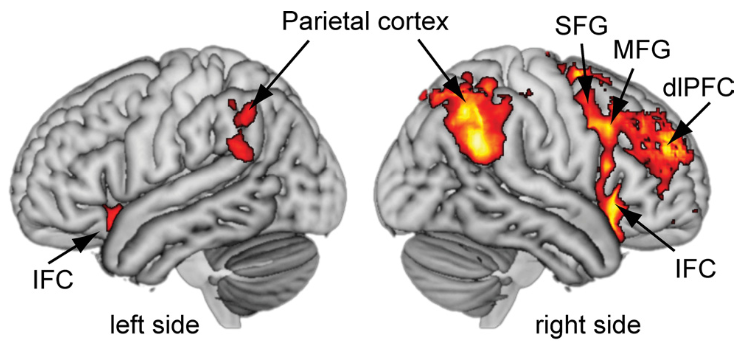} }} 
    \caption{An analogy between human cognition and LLMs. (a) and (b) human actions controlled consciously or unconsciously rely on partially distinct brain circuits. (a) Unconscious control in humans is maintained by a few specialised brain regions, such as the anterior insula and the presupplementary motor area (pre-SMA).  (b) while voluntary control engages a broader network, activating many regions within the parietal and prefrontal lobes \cite{van2010unconscious}. Unconscious control is typically fast and instinctive, often driven by automatic processes, whereas conscious control tends to involve more deliberate, computational, and in-depth thinking, allowing for careful reflection and thorough analysis.
    }%
    \label{fig:consciouscontrol}%
\end{figure}

Interestingly, in human cognition, two correlated yet distinct modes of cognitive processing are presented to guide human decision-making and behaviours \cite{kahneman2011thinking}, each of which has partially distinction brain circuits and neural pathways ( Fig.~\ref{fig:consciouscontrol} and also see \cite{van2010unconscious}). System 1 thinking is fast, automatic, and intuitive, operating effortlessly and often unconsciously. It relies on neural pathways that enable rapid processing, especially in situations needing quick reactions or when cognitive resources are constrained. System 2 thinking is deliberate, effortful, and conscious, involving focused attention and analytical reasoning. It processes information more slowly and is used for complex problem-solving, logical reasoning, and decision-making tasks.
o1 is an exciting development for AI, as LLMs can now not only generate rapid responses using learned patterns but, more significantly, simulate complex reasoning processes through mechanisms like chain of thought or other forms of search, similar to how humans engage in deeper, step-by-step thinking\footnote{It is important to note that incorporating chain-of-thought processes in AI does not imply human-like consciousness. Instead, these mechanisms enhance reasoning and problem-solving by breaking tasks into manageable steps without suggesting any form of self-awareness or subjective experience.}.

ChatGPT o1's improved reasoning skills have many implications for multiple fields, including science, coding, and mathematics. In coding competitions, a specialised version of o1 achieved impressive results, scoring in the 49th percentile in the 2024 International Olympiad in Informatics and outperforming 93\% of human competitors in simulated Codeforces contests. Beyond its technical capabilities, o1 also represents progress in AI safety and alignment. The model's chain of thought reasoning provides new opportunities for integrating human values and principles, resulting in improved performance on safety evaluations and jailbreak tests.

The idea of chain of thought reasoning and step-by-step thinking in Large Language Models (LLMs) is not new. Previous research has shown that simply adding instructions like ``describe your reasoning in steps" or ``explain your answer step by step" to the input questions or providing few shot examples can trigger LLMs to generate intermediate reasoning steps (as illustrated in Fig.~\ref{fig:cot}) and subsequently improve problem-solving, especially in tasks like math and coding \cite{wei2022chain, nye2021show}. However, these approaches build on existing LLMs without truly embedding the chain of thought ability within the models themselves. As a result, LLMs cannot inherently learn this reasoning capability, leading to active research on how to integrate it directly into model training. Proposed methods range from collecting specialised training data to building reward models \cite{ouyang2022training,li2022making,luo2024improve} and increasing the computational complexity of decoding \cite{snell2024scaling,wu2024empirical}, but none have yet achieved significant performance breakthroughs at scale.

It remains unclear whether OpenAI's o1 innovation is rooted in the model itself, rather than relying on external prompting systems. If it indeed involves explicitly embedding step-by-step reasoning natively within the architecture, this would represent a significant breakthrough.  Building on substantial performance gains, OpenAI o1 has shown that the scaling principles traditionally applied during training \cite{kaplan2020scaling,snell2024scaling} are now relevant to the inference phase. We should reallocate our computational focus, balancing pre-training efforts with efficient use of inference-time computation. Allowing LLMs to enhance their outputs with increased test-time computing is an essential step towards creating generally self-improving agents capable of managing open-ended strong reasoning and decision-making tasks. This direction,  which we refer to as LLM-Native Chain-of-Thought (NativeCoT), should be able to inherently mirror the deliberate, analytical process possessed by human's System 2 thinking \cite{kahneman2011thinking}.


Given that o1 is a closed-source system, the precise techniques used to achieve such strong reasoning capabilities remain largely a mystery. In this article, we will provide a comprehensive overview of the relevant literature and offer insights into what we believe are the core techniques and methods underpinning this breakthrough.  Additionally, we will propose our ideas for implementing an open-source counterpart, which could accelerate research in this area. Our proposals will draw inspiration from recent work, including ours on data acquisition, reinforcement learning based training, and search and MCTS-based decoding for improving reasoning capabilities in existing models.

In the next section, we will discuss two challenges commonly encountered by typical autoregressive LLMs, highlighting the need for a world model and a chain-of-thought mechanism. We will then present an MDP formulation for incorporating native CoT within LLMs (resulting in o1-like reasoning models) and explore its implementation details. Finally, we conclude with bibliographic remarks and suggest future research directions.

\section{The Challenges with Autoregressive LLMs}

Autoregressive language models (LLMs) generate sequences of text by predicting the next token (e.g., word) in the sequence given the previous tokens \cite{vaswani2017attention}. Mathematically, they are based on the principle of conditional probability. The task is to model the joint probability of a sequence of tokens \(\mathbf{x} = (x_1, x_2, \dots, x_T)\), where \(T\) is the length of the sequence, by factorising it into a product of conditional probabilities using the chain rule of probability. 


Given a sequence of tokens \(\mathbf{x} = (x_1, x_2, \dots, x_T)\), an autoregressive language model estimates the joint probability \(P(\mathbf{x})\) as:
\[
P(\mathbf{x}) = P(x_1, x_2, \dots, x_T) = \prod_{t=1}^{T} P(x_t \mid x_1, x_2, \dots, x_{t-1}),
\]
where the model predicts the probability of each token \(x_t\) based on all preceding tokens in the sequence \(x_1, x_2, \dots, x_{t-1}\). Typically, this is achieved using neural networks like transformers \cite{vaswani2017attention}, which are trained to minimise the negative log-likelihood of the training data. For an explanation of the training steps, please refer to Appendix A.

At inference time, the model generates text by typically sampling tokens sequentially from the probability distribution \(P(x_t \mid x_1, x_2, \dots, x_{t-1})\) until a stop token is reached or a predefined maximum length is achieved. The model works as follows:
Firstly, start with a given sequence or a start token (if generating from scratch).
Secondly, at each step \(t\), predict the next token \(x_t\) based on the previously generated tokens \((x_1, x_2, \dots, x_{t-1})\).
At last, continue sampling until the sequence is complete. For a simple three-token sequence \(\mathbf{x} = (x_1, x_2, x_3)\), the probability of the sequence would be:
\[
P(\mathbf{x}) = P(x_1) \cdot P(x_2 \mid x_1) \cdot P(x_3 \mid x_1, x_2).
\]

This formulation underpins the operation of autoregressive LLMs like GPT-style models.
The learning is achieved by minimising mistakes in predicting subsequent tokens (words). The first challenges is this \emph{predicting next tokens} objective. While some people propose that predicting next tokens might pave the way for general intelligence (AGI), we intend to argue is that solely focusing on predicting the next word caps the potential for intelligence. A different optimisation target and learning paradigm might be necessary to foster deeper intelligence.

To illustrate the limitations of purely predictive models, let's consider the domain of chess mastery. In this context, each chess move can be conceptualised as a token, with a complete chess representing a "sentence" in the "language of chess" - a sequence of moves from the opening to the endgame. Suppose we have access to an extensive dataset of chess games, but all from players with Elo ratings below 2000 (a standardised measure of player skill) \cite{elo1978rating}.
If we train a chess agent solely by minimising token prediction errors based on these games, we would likely constrain the agent's performance to within the ability range of these sub-2000 Elo players. This approach would essentially optimise the agent towards emulating the average or typical play of these players, potentially incorporating their mistakes and suboptimal strategies.
This phenomenon can be characterised as what we called an "intelligence upper bound," a concept that can be rigorously derived from recent research in offline reinforcement learning and imitation learning \cite{levine2020offline}. The agent, in this case, is limited by the quality of the demonstrations it learns from, unable to surpass the skill level present in its training data. This limitation underscores a crucial challenge in AI development: how to enable systems to transcend the boundaries of their training data and develop novel, potentially superior strategies. 

Conversely, when data is leveraged to develop a deeper understanding, or a \emph{world model}, of chess dynamics, it may pave the way for the evolution of sophisticated strategies and tactics that go beyond mere imitation of behaviours observed in the training data. A world model presents the agent's understanding of the environment, in this case, the chess rules, i.e., how a move would change the status of the game and what the winning chance of a given move is.  Learning and refining this world model, coupled with the ability to simulate potential outcomes, could potentially empower an AI agent to surpass the 2000 Elo benchmark. The simulation capabilities afforded by these internal world models would enable deep thinking (simulation), thereby enhancing the agent's reasoning and generalisation capabilities. Model-based strategies like Monte Carlo Tree Search (MCTS) serve as classic illustrations of this approach \cite{silver2017mastering}. The transition to System 2 type reasoning, as potentially exemplified by ChatGPT o1, likely relies on establishing a certain type of World Model and utilising reinforcement learning (reward maximisation) rather than solely minimising prediction errors. This shift in approach may be one of the key transitional techniques behind ChatGPT o1's enhanced reasoning capabilities.

By combining the predictive power of large language models with the strategic depth of reinforcement learning and World Modelling, AI systems like o1 can potentially engage in more sophisticated problem-solving and decision-making processes. This hybrid approach allows for both rapid pattern recognition (akin to System 1 thinking) and deliberate, step-by-step reasoning (characteristic of System 2 thinking), potentially explaining the significant leap in performance observed in o1.

The second challenge, from a computational complexity perspective, is that Large Language Models (LLMs) inherently operate within the constraints of quadratic computational complexity \cite{lin2020limitations}. This limitation becomes particularly apparent when LLMs encounter multi-step mathematical challenges. However, the "chain of thoughts" concept offers a potential mitigation to this constraint \cite{wei2022chain}.  It extends responses through a series of "thought" outputs, therefore allows a certain amount of additional computation resources; it essentially acts as a \emph{limited memory} that supports writing but lacks the capacity for deletion or overwriting. While this approach has shown promise, it still falls short of a fully dynamic memory system and is not natively incorporated into the decoding stage. This necessity underscores the demand for advanced computational architectures that transcend the capabilities of current transformer decoder networks. Indeed, there is a need to implement sophisticated model-based strategies akin to Monte Carlo Tree Search (MCTS) witnin the inference and decoding stage \cite{feng2023alphazero}. 

Such an advanced inference-time computation system would enable AI models to maintain and dynamically update a representation of the problem space, facilitating more complex reasoning processes. This approach \cite{christianos2023pangu} aligns with the concept of working memory in cognitive science, which is crucial for complex problem-solving and deliberative thinking. By integrating these capabilities, AI systems could potentially simulate multiple steps ahead, evaluate different scenarios, and make more informed decisions --- mirroring the deliberative processes observed in human expert reasoning.

\begin{figure*}[t]
    \centering
    \includegraphics[width=1\linewidth]{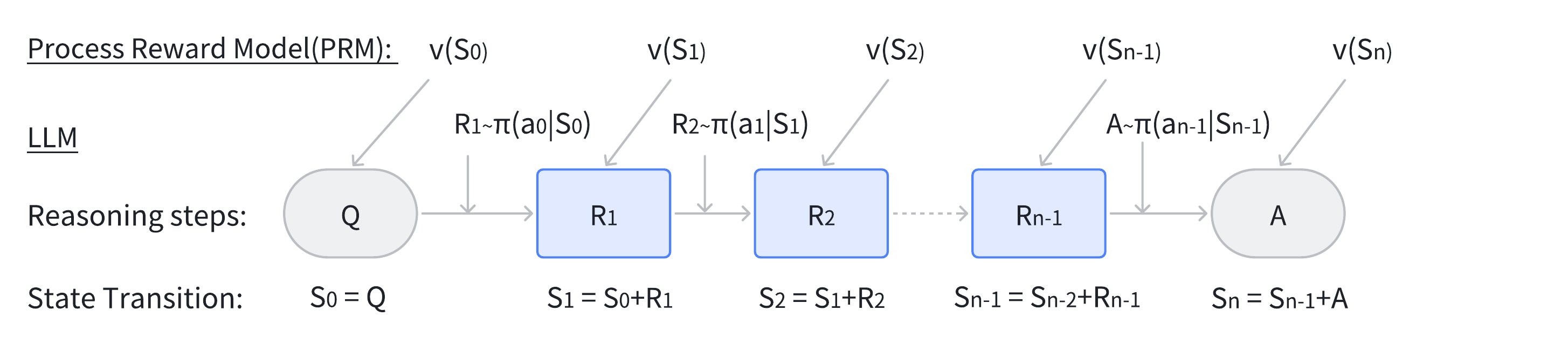}
    \caption{In this MDP formulation, the LLM is tasked with generating reasoning steps and the final answer to a question in a step-by-step manner. The LLM policy operates by generating tokens, which form higher-level reasoning constructs. The states represent the sequence of reasoning steps so far, and actions correspond to the selection of new reasoning steps or the final answer. The LLM policy governs the choice of actions, and the process-reward model (PRM) provides feedback on the quality of reasoning steps and the final answer. By optimising the policy to maximise the reward, the LLM can be guided by PRM to generate accurate and meaningful reasoning processes.
}
    \label{fig:main_fig}
\end{figure*}

\section{LLM Reasoning as a Markov Decision Process}

To model the process of reasoning in tasks such as question answering or problem solving, we structure the reasoning task using the Q → \{R\} → A sequence, where:
\begin{itemize}
    \item \textbf{Q:} Represents the question or prompt that initiates the reasoning process.
    \item \textbf{R:} Represents the sequence of intermediate reasoning steps the model generates to build toward the solution.
    \item \textbf{A:} Represents the final answer or solution produced after the reasoning steps.
\end{itemize}
This structure allows the LLM to generate a sequence of reasoning steps that logically connect the question \( Q \) to the final answer \( A \).

We can define the reasoning process as a Markov Decision Process (MDP) \cite{bellman1958dynamic}.  A MDP representation offers a flexible framework for modelling reasoning. It allows the model to autoregressively generate sequential reasoning steps toward the final answer, while also enabling a tree structure by sampling multiple paths at each step for alternative reasoning trajectories. By combining both approaches-sequential and branching reasoning-the model can explore diverse solutions, creating a versatile and comprehensive reasoning process.

We are now ready to describe the reasoning process in terms of states, actions, policies, and rewards, where the LLM's task is to incrementally generate a coherent sequence of tokens that correspond to reasoning steps and the final answer.

The state \( s_t \) at timestep \( t \) represents the current state of the reasoning process, including the question and the reasoning steps generated so far. Formally, the state is defined as:
\[
s_t = (Q, R_1, \dots, R_{t-1}),
\]
where \( Q \) is the initial question or prompt, and \( R_1, \dots, R_{t-1} \) are the reasoning steps generated up to timestep \( t \). The initial state \( s_0 \) contains just the question:
\[
s_0 = Q.
\]

As reasoning progresses, the intermediate states include both the question and the reasoning steps generated so far. The process continues until the final answer is generated.

An action \( a_t \in A \) at timestep \( t \) corresponds to the selection of the next reasoning step or the final answer. The action space \( A \) consists of two types of actions:
\begin{itemize}
    \item \textbf{Reasoning Step (R):} The action selects a reasoning step \( R_t \) to append to the current state.
    \item \textbf{Final Answer (A):} The action selects the final answer \( A \), which concludes the reasoning process.
\end{itemize}
For intermediate steps, the action is:
\[
a_t = R_t,
\]
and the new state becomes:
\[
s_{t+1} = s_t + R_t.
\]
For the final step, the action selects the final answer:
\[
a_T = A,
\]
and the final state becomes:
\[
s_T = s_{T-1} + A.
\]

The policy \( \pi \) defines the strategy the model uses to choose the next action (i.e., reasoning step or final answer) given the current state. The policy is essentially the LLM, learned during training and represents the probability distribution over possible reasoning steps or the final answer, conditioned on the tokens generated so far:
\[
\pi_{LLM}(a_t \mid s_t) = P(a_t \mid Q, R_1, \dots, R_{t-1}).
\]

At each timestep, the model uses this policy to select the next action based on the current state, incrementally building towards the final answer.

Given the autoregressive nature of the LLM, the transition from one state to the next is deterministic and also given. The next state \( s_{t+1} \) is fully determined by appending the selected action \( a_t \) (a reasoning step or the final answer) to the current state \( s_t \). Therefore, the transition function is:
    \[
    s_{t+1} = s_t + a_t.
    \]
    
This means that once a reasoning step \( R_t \) or final answer \( A \) is selected, the state \( s_{t+1} \) is uniquely defined by concatenating this action to the existing sequence of tokens.

The reward provides feedback on the quality of the generated reasoning steps and the final answer. In this context, the reward is obtained as the model generates reasoning steps and the final answer. The rewards can be defined as:
\begin{itemize}
    \item \textbf{Intermediate Reward:} For generating correct or meaningful reasoning steps, intermediate rewards are assigned positive values. Incorrect or irrelevant steps may yield negative rewards.
    \item \textbf{Final Reward:} The largest reward is given when the model generates the correct final answer \( A \), completing the reasoning process.
\end{itemize}
Thus, the reward at each timestep \( t \) is:
\[
v_t = v(\text{R}_t \mid Q, R_1, \dots, R_{t-1}),
\]
and for the final step:
\[
v_T = v(A \mid Q, R_1, \dots, R_n).
\]

The model learns to optimise its policy to maximise the cumulative expected reward over the entire reasoning process.

\textbf{Relationship Between Token Generation and Reasoning} The LLM operates at two levels simultaneously: the level of token generation and the level of reasoning steps and final answers. At the most granular level, the LLM generates tokens autoregressively, meaning it generates one token at a time, conditioned on the previously generated tokens:
\[
P(x_t \mid x_1, x_2, \dots, x_{t-1}).
\]

At each timestep \( t \), the LLM generates a token \( x_t \) from its vocabulary based on the context provided by previous tokens. These tokens form higher-level constructs such as reasoning steps \( R_t \) and the final answer \( A \).

\begin{itemize}
    \item \textbf{Reasoning Steps (R):} Each reasoning step \( R_t \) is composed of a sequence of tokens \( \{x_{t_1}, x_{t_2}, \dots, x_{t_k}\} \) generated by the LLM. These tokens represent a coherent step in the reasoning process, such as a logical deduction or intermediate conclusion.
    \item \textbf{Final Answer (A):} The final answer \( A \) is similarly composed of a sequence of tokens that form the solution or response to the question. Once the LLM has generated sufficient reasoning steps, it produces the final answer in an autoregressive manner, token by token.
\end{itemize}

We are now ready to a world model for LLMs exactly:
\newtheorem{definition}{Definition}
    

\begin{definition}[World Model of LLM]
A \emph{world model of LLM} is defined as \( (\mathcal{T},\mathcal{R} )\), where:
\begin{itemize}
    \item The transition model \( \mathcal{T}(s_t, a_t) \) is deterministic as the next state \( s_{t+1} \) is uniquely defined by the current state \( s_t \) and the action \( a_t \) (i.e., the generated token or reasoning step), so:
    \[
    s_{t+1} = s_t + a_t.
    \]
    \item \( \mathcal{V}(s_t, a_t) \) is the process-reward model (PRM) that evaluates the quality of the action \( a_t \) taken in state \( s_t \). It reflects how appropriate or effective the generated reasoning step or token is in progressing towards the final answer:
    \[
    \mathcal{V}(s_t, a_t) = v_t.
    \]
\end{itemize}
Since the transition is deterministic and follows directly from the policy, the process-reward model (PRM) \( \mathcal{R}(s_t, a_t) \) encapsulates the entire interaction between the LLM and its environment, evaluating how well each reasoning step or token contributes to reaching the final answer.
\end{definition}

\section{Practical Implementation}
Next, we examine how to collect the intermediate reasoning data, use it to train the process-reward model (PRM), leverage the PRM to train the LLM policy, and guide the reasoning process during the decoding phase.

\begin{figure*}[t]
    \centering
    \includegraphics[width=1\linewidth]{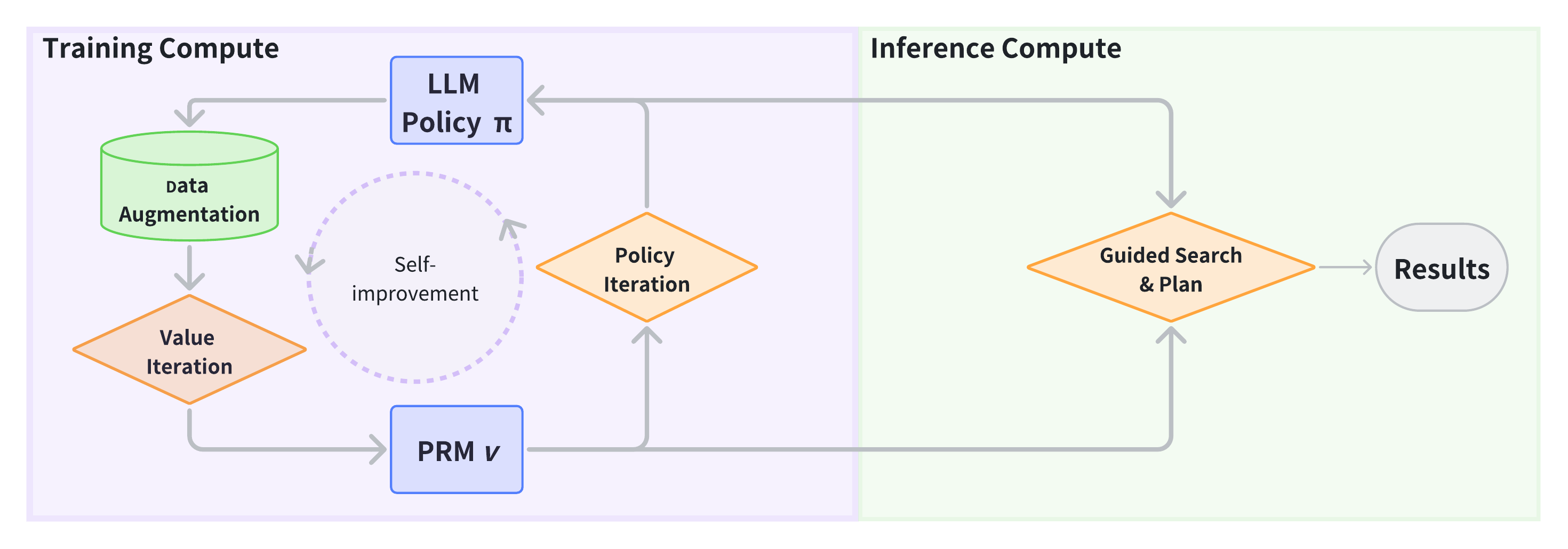}
    \caption{Combining the value function from the PRM with the LLM's policy generation ensures guided and controlled results. During training, the generation produced by the LLM's policy and the evaluation provided by the PRM reinforce each other, leading to continuous self-improvement and refinement of both components.}
    \label{fig:interactionprmllm}
\end{figure*}

\subsection{Automatic Acquisition of Reasoning Steps Data}
As discussed, we require reasoning trajectories to stimulate advanced reasoning while covering a wide range of tasks.  For fine-tuning a LLM, we typically  have \{Q and A\} pairs, however lacking the ground-truth of underlying reasoning steps \{R\}: 
\[
\begin{array}{c}
\textbf{Question Q} \\
\downarrow \\
\quad\quad\quad\quad\quad Reasoning Step 1:\quad  r_1 \quad (\text{Reward}_1) \\
\downarrow \\
\quad\quad\quad\quad\quad Reasoning Step 2:\quad r_2 \quad (\text{Reward}_2) \\
\downarrow \\
...\\
\downarrow \\
\quad\quad\quad\quad\quad\quad\quad\quad\quad\quad \textbf{Answer A}:  \quad r_A\quad (\text{Final Reward})
\end{array}
\]


A straightforward approach would be to label the reasoning steps manually by humans \cite{uesato2022solving,lightman2023let}. 
However, a particularly effective method for collecting data and improving LLM reasoning without requiring human supervision is the Self-Taught Reasoner (STaR) technique \cite{zelikman2022star}, among others. In this approach, the model generates intermediate reasoning steps autonomously and uses them to validate its internal reasoning capabilities. This method builds on the ability of LLMs to reason from a question \( Q \) to a final answer \( A \), by generating intermediate steps \( \{R_1, R_2, \dots, R_n\} \) and verifying their correctness using the model’s own policy. Namely, the method begins by employing the LLM's policy (may add few shot prompts), denoted \( \pi_{\text{LLM}} \), to generate reasoning steps \( \{R\} \) conditioned on the initial question \( Q \) and final answer \( A \). This generation can be expressed as follows:
\[
\{R\} \thicksim \pi_{\text{LLM}}(\cdot \mid Q, A),
\]
where the LLM produces a sequence of intermediate reasoning steps \( \{R_1, R_2, \dots, R_n\} \) that aim to logically connect the question \( Q \) to the correct final answer \( A \). These steps serve as a form of internal decomposition of the reasoning task, which is crucial for complex multi-step problems where direct question-answer pairs may be insufficient for training the model to reason effectively.

Once the intermediate reasoning steps \( \{R\} \) are generated, the next phase involves verifying their correctness. This is achieved by using the LLM’s policy again to check whether the reasoning steps, when combined with the original question \( Q \), lead to the correct answer \( A \). Formally, this verification step is represented by:
\[
A' \thicksim \pi_{\text{LLM}}(\cdot \mid Q, \{R\}),
\]
where, \( A' \) is the model's prediction of the answer based on the question \( Q \) and the generated reasoning steps \( \{R\} \). If \( A' \) matches the original correct answer \( A \), then the reasoning steps \( \{R\} \) are considered valid. Thus, the correctness of \( \{R\} \) is determined by the condition: $A' \thickapprox A$. This self-validation mechanism enables the model to autonomously identify correct reasoning steps, reinforcing its internal logical consistency without external feedback.

The collected new reasoning steps \({\{Q, \{R\},A\}} \) can be used  to further train the LLM's policy \( \pi_{\text{LLM}} \), reinforcing the generation of effective reasoning steps. This iterative process can be expressed as:
\[
\pi_{\text{LLM}} \leftarrow \pi_{\text{LLM}} + \text{feedback from } \{Q, \{R\},A \}.
\]
For longer reasoning sequences, techniques such as Monte Carlo Tree Search (MCTS) \cite{feng2023alphazero,luo2024improve} are employed to guide the LLM policy to find correct reasoning steps efficiently in a more fine-grained manner. These tree-based methods help in finding optimal reasoning paths by exploring various possibilities and simulating multiple outcomes in each reasoning stage. This is particularly useful for complex tasks like math problem-solving and agent-based decision-making, where intermediate steps have multiple paths.

\subsection{Self-reinforced Training}
As illustrated in Fig.~\ref{fig:interactionprmllm}, the PRM $v(s)$ and LLM policy ($\pi_{LLM}$) can be mutually reinforced to improve themselves, which will be explained next.

\subsubsection{Value Iteration for PRM}
Once the reasoning data has been collected, the next step is to train the world model, also referred to as the Process-Reward Model (PRM), i.e., since the state transitions are deterministic and known, the focus shifts to learning a general reward model that can later be used to guide the search, reasoning, and decoding processes. This reward model, often called the verifier, denoted as \( v_{\text{PRM}}(s) \), can be trained using a dataset of annotated reasoning steps. The training typically involves optimising a classification loss function based on the correctness of the reasoning steps \cite{luo2024improve}:
\[
\mathcal{L}_{\text{PRM}} = \sum_{i=1}^{N} \left[ \hat{v}_i \log v_i + (1 - \hat{v}_i) \log (1 - v_i) \right],
\]
where \(v_i=r_i\) represents the correctness label for the \(i\)-th example step, indicating whether the reasoning process for that example is correct. The verifier’s prediction, \(\hat{v}_i(s) \), is the score output by the PRM for the state \(s\), representing the reward for the reasoning step or the final answer. Since this is a classification approach, there is no distinction between the reward for an intermediate step and the potential reward it could lead to and all the reasoning steps are assumed to be independent. The model simply evaluates whether the reasoning step or answer is correct at that point in the process, treating all rewards in a uniform manner without considering the future impact of intermediate steps.

However, an alternative approach involves viewing the PRM as a value function that can be trained via a value iteration method, enabling it to predict cumulative rewards and guide the reasoning process through optimal action selection \cite{feng2023alphazero}.  Consider a reasoning process where the state \( s \) represents the current reasoning state, incorporating all previous reasoning steps. The objective of the value iteration method is to learn a value function \( V_{\theta}(s) \), parameterised by \( \theta \), that predicts the expected cumulative reward starting from state \( s \). This value function guides the reasoning process by evaluating the potential outcomes of different actions. 
 \( r_{\phi}(s) \) – the reward function, which assigns a scalar reward to state \( s \) based on the correctness of intermediate reasoning steps or the final answer. \( \gamma \) is the discount factor, which determines the relative importance of future rewards. The Bellman equation \cite{bellman1958dynamic} for the PRM is:
\[
V_{\theta}(s) = r(s) + \gamma \max_{a} V_{\theta}(a+s),
\]
where \( s'=a+s \) is the next state reached by taking action \( a \) in state \( s \). The reward function \( r(s) \) can be sparse, providing rewards only for correct conclusions, or dense, providing partial rewards for intermediate steps. We define the TD loss function for learning the parameters \( \theta \) of the value function as the squared error between the current value and the Bellman target:
    \[
    L(\theta) = \sum_{i=1}^{N} \left( V_{\theta}(s_i) - \left[ r(s_i) + \gamma \max_{a} V_{\theta}(s_i+a) \right] \right)^2.
    \]
    
We can then obtain the parameters \( \theta \) of the value function by minimising the loss \( L(\theta) \) using gradient descent or another optimisation technique.

\subsubsection{Policy Iteration for LLM Policy}
Once PRM obtained, one can train the LLM policy for enhanced reasoning. This requires methodologies that go beyond traditional supervised learning frameworks. PRM plays an essential role in this process by incorporating online reinforcement learning to optimise reasoning tasks \cite{ouyang2022training}. However, a typical RLHF work such as \cite{ouyang2022training} can be used but may not be ideal for large language model training. 

Let us look at Group Relative Policy Optimisation (GRPO) \cite{shao2024deepseekmath}. We assume that for each question \( Q=q \), the policy generates reasoning steps \( \{o_1, o_2, \dots, o_G\} \), and each output \( o_i \) consists of multiple steps \( \{a_{i,1}, a_{i,2}, \dots, a_{i,K_i}\} \), where \( K_i \) is the total number of reasoning steps (or tokens) in output \( o_i \). We slightly abuse our previous notation by using $o$ to represent all outputs, including both reasoning steps and final answers. We can now formulate the GRPO optimisation for learning the LLM policy via the PRM as follows.

For each question \( q \), GRPO samples a group of outputs \( \{o_1, o_2, \dots, o_G\} \) from the old policy \( \pi_{\theta_{\text{old}}} \), and the goal is to optimise the policy by maximising the following objective:
\[
J_{\text{GRPO}}(\theta) = \mathbb{E}_{q \sim P(Q), \{o_i\}_{i=1}^{G} \sim \pi_{\theta_{\text{old}}}(O|q)} \left[ \frac{1}{G} \sum_{i=1}^{G} \frac{1}{K_i} \sum_{t=1}^{K_i} \min\left( \hat{\rho}_{i,t} A_{i,t}, \text{clip}\left( \hat{\rho}_{i,t}, 1 - \epsilon, 1 + \epsilon \right) A_{i,t} \right) - \beta D_{\text{KL}} \left( \pi_\theta \| \pi_{\theta_{\text{ref}}} \right) \right],
\]
where:
\begin{itemize}
    \item \( q \sim P(Q) \) denotes sampling a question \( q \) from a distribution of questions \( P(Q) \),
    \item \( \{o_i\}_{i=1}^{G} \sim \pi_{\theta_{\text{old}}}(O|q) \) represents the group of outputs sampled from the old policy \( \pi_{\theta_{\text{old}}} \),
    \item \( \hat{\rho}_{i,t} = \frac{\pi_\theta(a_{i,t}|q, o_{i,<t})}{\pi_{\theta_{\text{old}}}(a_{i,t}|q, o_{i,<t})} \) is the importance weight (probability ratio) for action \( a_{i,t} \) at step \( t \) in output \( o_i \),
    \item \( A_{i,t} \) is the advantage at reasoning step \( t \) of output \( o_i \), calculated based on relative rewards (see below),
    \item \( \epsilon \) is the clipping parameter that prevents excessive updates (as in PPO \cite{schulman2017proximal}),
    \item \( \beta \) is a hyperparameter controlling the strength of KL regularisation,
    \item \( D_{\text{KL}} \left( \pi_\theta \| \pi_{\theta_{\text{ref}}} \right) \) is the KL divergence between the trained policy \( \pi_\theta \) and a reference policy \( \pi_{\theta_{\text{ref}}} \), used as regularisation.
\end{itemize}

The advantage function \( A_{i,t} \) for the action \( a_{i,t} \) taken at step \( t \) in output \( o_i \) is calculated based on the rewards from both reasoning steps and the final step. The rewards are normalised using the rewards across all outputs in the group for a specific question. Let the normalised reward for step \( t \) of output \( o_i \) be:
\[
 \bar r_i^{(t)} = \frac{r_i^{(t)} - \text{mean}(R)}{\text{std}(R)},
\]
where \[
R = \left\{ \{r_1^{\text{index}(1)}, \dots, r_1^{\text{index}(K_1)}\}, \dots, \{r_G^{\text{index}(1)}, \dots, r_G^{\text{index}(K_G)}\} \right\},
\]
 represents the rewards from all reasoning steps across all outputs in the group \( G \), where \(\text{index}(j)\) is the end token index of the \(j\)-th step, and \(K_i\) is the total number of steps in the \(i\)-th output; and \( \text{mean}(R) \) and \( \text{std}(R) \) are the mean and standard deviation of the group rewards.

The advantage \( A_{i,t} \) for the \( t \)-th step of output \( o_i \) is the sum of normalised rewards from step \( t \) to the final step \( K_i \):
\[
A_{i,t} = \sum_{j=t}^{K_i} \bar r_i^{(j)},
\]
where \( \bar r_i^{(j)} \) is the normalised reward for reasoning step \( j \) in output \( o_i \). This advantage function encourages the model to optimise for both intermediate reasoning steps and the final step, by rewarding reasoning paths that yield higher relative performance within the group.

Rather than incorporating a KL penalty directly into the reward, GRPO regularises the policy by adding the KL divergence between the current policy \( \pi_\theta \) and a reference policy \( \pi_{\theta_{\text{ref}}} \) directly into the loss function. This ensures that the updated policy does not deviate excessively from the reference policy during training, helping maintain stability.


This GRPO formulation, specifically adapted for reasoning tasks with process reward models, optimises LLM policy by leveraging group relative rewards across reasoning steps and final steps. The normalised advantage function is computed based on relative performance, encouraging the policy to favours outputs that perform better within a group of sampled outputs. Additionally, KL regularisation ensures that the updated policy remains close to a reference policy, improving training stability and efficiency. This framework provides a robust approach for guiding LLM reasoning through PRM-based optimisation.

One can explore more efficient offline methods such as token-level DPO  \cite{zeng2024token} without PRM but with sequential reasoning data. For details, please refer to the paper.

\subsection{Inference-time Computation}
Once trained, the LLM policy must efficiently generate outputs during inference. Autoregressive generation—where tokens are predicted one by one based on previous tokens—is widely used in LLMs. However, for reasoning tasks, more sophisticated decoding techniques are necessary.

To strike a balance between efficiency and effectiveness, the work \cite{snell2024scaling,wu2024empirical} found that reasoning tasks benefit from more flexible approaches like beam search. In beam search, multiple possible sequences (or beams) are generated simultaneously, and the best candidate is chosen based on cumulative probability. For even more complex reasoning tasks, look ahead model such as MCTS is used. MCTS \cite{feng2023alphazero} simulates multiple reasoning paths and evaluates them based on a reward system, selecting the one with the highest expected reward. This allows the model to explore a wider range of possibilities during inference, increasing its chances of arriving at an optimal solution. With an MDP, we could formally define the reasoning process structure.

\begin{figure*}[t]
    \centering
    \includegraphics[width=1\linewidth]{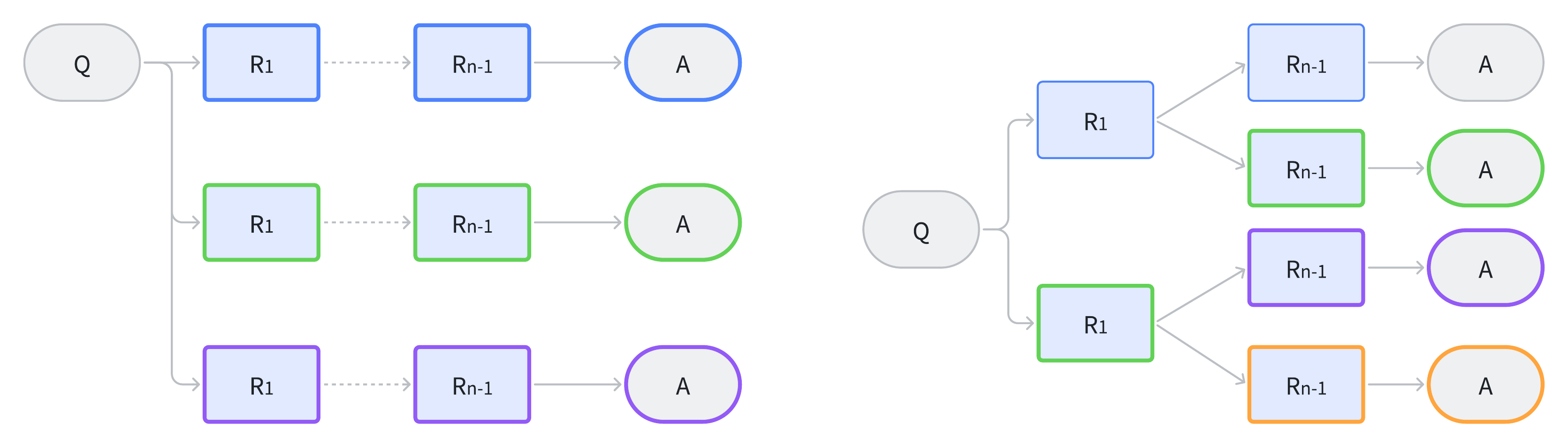}
    \caption{With the PRM, the LLM can perform non-autoregressive reasoning through three approaches: 1) sampling multiple reasoning trajectories, 2) performing a Monte Carlo search over a tree structure of potential reasoning paths, or 3) combining both methods to enhance flexibility and robustness in reasoning.}
    \label{fig:reasonpro}
\end{figure*}

\begin{definition}[Native Chain-of-Thought]
\emph{Native Chain-of-Thought (NCoT)} refers to the inherent reasoning capability of a large language model (LLM), which allows it to autonomously perform step-by-step, structured reasoning without external prompts. This capability is formalised as a Markov Decision Process (MDP) \( (\mathcal{S}, \mathcal{A}, \pi, \mathcal{R}) \), where:

\begin{itemize}
    \item \( \mathcal{S} \) is the state space, representing the sequence of tokens or reasoning steps generated up to a given point.
    \item \( \mathcal{A} \) is the action space, which consists of potential reasoning steps \( R_t \) or the final answer \( A \).
    \item \( \pi_{LLM}(a_t \mid s_t) \) is the policy (also the LLM) that governs the selection of actions, determining the next reasoning step or final answer based on the current state \( s_t \).
    \item \( \mathcal{R}(s_t, a_t) \) is the process-reward model (PRM), which assigns a reward \( r_t \) based on the quality and relevance of the selected action \( a_t \), guiding the reasoning process.
\end{itemize}

The model can either follow a sequential reasoning path by unrolling the MDP or explore multiple trajectories by sampling different reasoning steps at each state, forming a tree-like structure (Fig.~\ref{fig:reasonpro}). The process-reward model \( \mathcal{R} \) provides a guided search over this space, controlling the reasoning trajectory by favouring actions that lead to more meaningful or correct reasoning steps.
\end{definition}

\section{Bibliographic Remarks}
\begin{figure*}[t]
    \centering
    \includegraphics[width=1\linewidth]{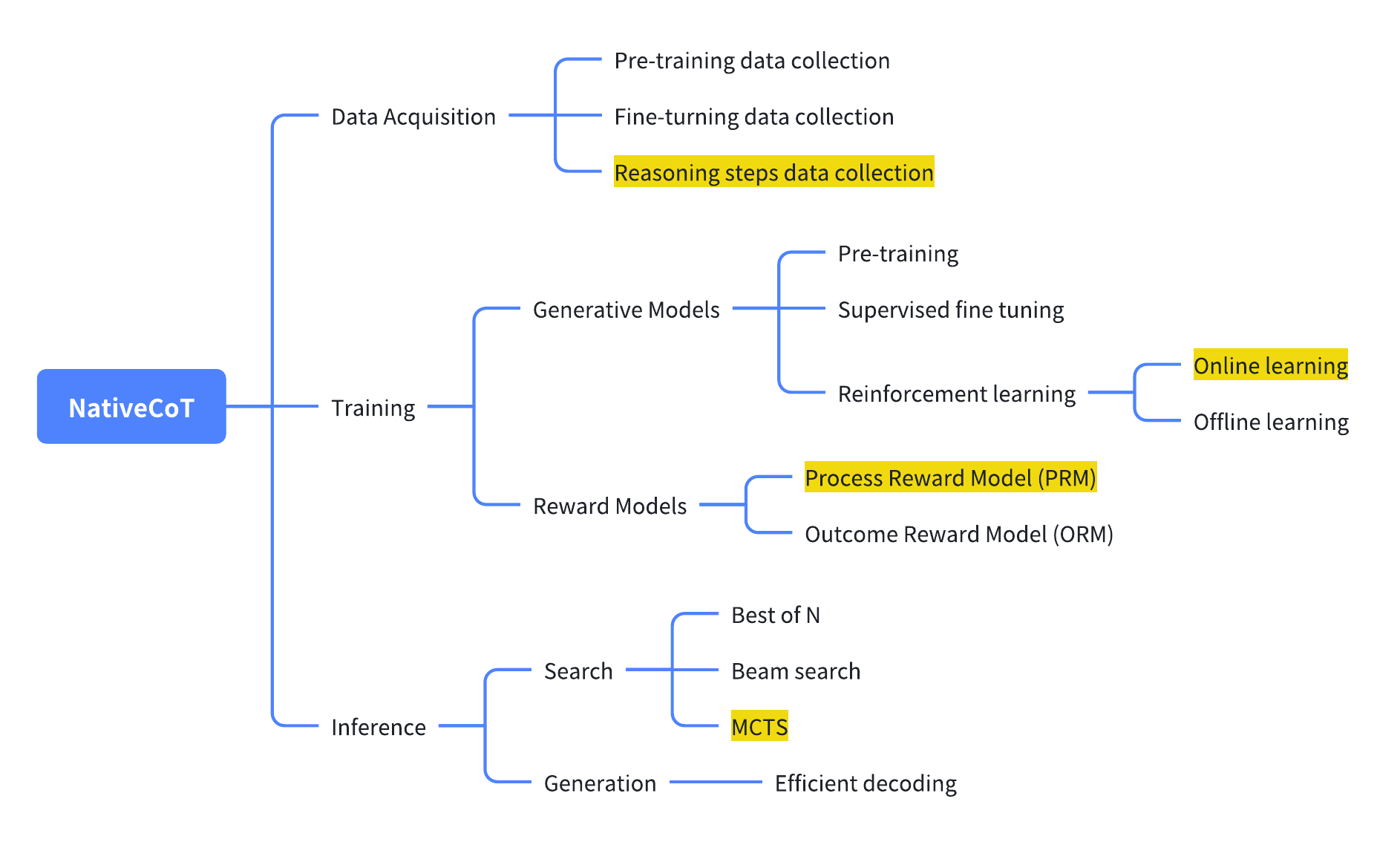}
    \caption{Research on LLM-native chain of thought. }
    \label{fig:landscape}
\end{figure*}

In the literature, significant attention has been given to inference-time computation, verifiers (also known as reward models), and data acquisition methods, all of which play a critical role in enhancing the reasoning capabilities of these models. In this section, we review and discuss several key papers in these areas, examining their contributions and limitations. The connection between these works and the broader research landscape is depicted in Fig.~\ref{fig:landscape}.

\subsection{Inference-Time Computing}

Several papers have focused on optimising LLM reasoning through inference-time computing. For instance, the paper \cite{feng2023alphazero} introduces a method that integrates Monte Carlo Tree Search (MCTS) with LLM decoding, a combination that has proven highly effective in guiding reasoning, particularly for complex, multi-step tasks. The inclusion of MCTS facilitates better decision-making by simulating potential future actions, enhancing the model's ability to plan its next steps. Similarly, the paper \cite{snell2024scaling} emphasises the importance of optimising test-time computation, empirically showing that inference-time reasoning enhancements can often yield more substantial improvements than simply scaling model parameters. This reflects a growing understanding that more compute during inference can be leveraged for higher quality reasoning without necessarily increasing the model's size.

Another approach is presented in \cite{goyal2023think}, which suggests using pause tokens to force models to pause and ``think" during reasoning. This method introduces an implicit reasoning model, encouraging the LLM to process information in chunks, mimicking human-like deliberation. 

\subsection{Verifier Models}

Verifier models (outcome-reward models and process-reward models) have become an important area of research in improving LLM reasoning reliability. Papers like \cite{cobbe2021training} introduced the earliest formal attempt (outcome reward only) at using verifiers in mathematical reasoning tasks, laying the groundwork for subsequent research. The follow-up work \cite{uesato2022solving} expands on the concept of verifiers, integrating process-based reasoning mechanisms, and was followed by OpenAI’s work on Process Reward Models (PRMs) \cite{lightman2023let}. These verifiers play a crucial role in ensuring the correctness of multi-step reasoning, addressing one of the major challenges in LLMs—maintaining coherence and accuracy over extended reasoning sequences.

A more recent addition to this line of research is \cite{li2022making}, which combines verifier models with majority voting to produce more reliable outputs in reasoning tasks. This method enhances the robustness of the verification process by cross-checking multiple reasoning paths and filtering out incorrect steps. Such advancements highlight the growing importance of verifiers in maintaining the accuracy of LLMs as they tackle increasingly complex reasoning challenges.

\subsection{Data Acquisition for Reasoning Tasks}

The acquisition of reasoning data has been another area of focus, particularly in papers like \cite{zelikman2022star}, which explores methods for automatically obtaining data related to reasoning steps. STaR introduces a self-teaching paradigm where the model improves its reasoning capabilities by generating and critiquing its own steps, leading to more reliable intermediate steps. The paper \cite{wang2024math} takes this approach further, showing how LLMs can be trained step-by-step without the need for costly human annotations, providing a more scalable solution to the reasoning data problem.

The work in \cite{wang2024multi} highlights the importance of practical data acquisition for reasoning tasks, particularly in coding problems. MCTS has been used for acquiring data in \cite{feng2023alphazero}, whereas it has been extended with linear search for efficiency in \cite{luo2024improve}. 

These papers suggest that for LLMs to advance in reasoning, innovative data acquisition methods, such as self-supervised learning and verification mechanisms, are essential to reduce the dependency on extensive human-labelled datasets.

\subsection{Understanding and System-Level Improvements}

Finally, there is a growing body of research aimed at understanding the mechanisms behind step-by-step reasoning in LLMs \cite{tutunov2023can,prystawski2024think}. The work in \cite{touvron2023llama} focused its analysis from graphical models for the chain of thought mechanism.  The paper \cite{prystawski2024think} explores the intrinsic reasons why reasoning emerges as a natural capability in LLMs. It suggests that reasoning is a byproduct of the way language models process localised experiences and knowledge. The paper \cite{luo2023critique} provides an empirical evaluation of LLMs' ability to critique their own reasoning, showing that self-critique is often limited, and this capability often emerges only when models are sufficiently large.

From a system perspective, the pangu-agent paper \cite{christianos2023pangu} introduces structured reasoning mechanisms beyond traditional models like OpenAI’s o1 model. This research reflects a shift toward more generalised reasoning agents that can handle a wider array of tasks with greater precision and flexibility, providing a vision of the next generation of reasoning models.





\bibliographystyle{abbrv}
\bibliography{references}

\appendix

\section{Standard Training Pipelines of LLMs}
The training procedure for LLM typically involves several stages, each building upon the previous one. In the pre-training stage, the model is trained on a massive online corpus using an autoregressive language modelling objective. The goal is to predict the next token given the previous tokens. For a given sequence of tokens \( \{x_1, x_2, \dots, x_T\} \), the token-level cross-entropy loss sums the negative log-probabilities of the true tokens at each position:
\[
\mathcal{L}_{\text{pretrain}} = -\sum_{t=1}^T \log P(x_t | x_{<t}; \theta),
\]
where $x_t$ is the $t$-th token, $x_{<t}$ represents all tokens before $t$, $\theta$ are the model parameters, and $P$ is the probability distribution over the vocabulary \cite{brown2020language}. \( p(x_t \mid x_{<t}) \) is the probability of the true token \( x_t \) given all previous tokens \( x_{<t} \). This loss measures how well the model predicts each token in the sequence.

After pre-training, the model is then fine-tuned on collected additional \{Question, Answer\} pairs. The objective is to maximise the likelihood of the correct answer given the question:
\[
\mathcal{L}_{\text{finetune}} = -\sum_{i=1}^N \log P(A_i | Q_i; \theta),
\]
where $Q_i$ and $A_i$ are the $i$-th question and answer pair, respectively \cite{raffel2020exploring}. 

Next, Reinforcement Learning from Human Feedback (RLHF) \cite{ouyang2022training} is then applied to further improve the model's instruction-following ability. This involves constructing a reward model $R(x)$ (by pair-wise training data) that estimates the quality of the model's outputs. The policy (language model) is then optimised using methods like Proximal Policy Optimisation (PPO) \cite{schulman2017proximal}:
\[
\mathcal{L}_{\text{RLHF}} = \mathbb{E}[R(Q,A)] - \beta \cdot \text{KL}(\pi_\theta (Q|A) \| \pi_{\theta_{\text{old}}}(Q|A)),
\]
where $\pi_\theta$ is the current policy, $\pi_{\theta_{\text{old}}}$ is the old policy, and $\beta$ is a hyperparameter controlling the strength of the KL divergence penalty.


\end{document}